# Quantitative Phase Imaging and Artificial Intelligence: A Review

YoungJu Jo, Hyungjoo Cho, Sang Yun Lee, Gunho Choi, Geon Kim, Hyun-seok Min, and YongKeun Park

(*Invited Paper*)

*Abstract*— **Recent advances in quantitative phase imaging (QPI) and artificial intelligence (AI) have opened up the possibility of an exciting frontier. The fast and label-free nature of QPI enables the rapid generation of large-scale and uniform-quality imaging data in two, three, and four dimensions. Subsequently, the AI-assisted interrogation of QPI data using data-driven machine learning techniques results in a variety of biomedical applications. Also, machine learning enhances QPI itself. Herein, we review the synergy between QPI and machine learning with a particular focus on deep learning. Further, we provide practical guidelines and perspectives for further development.**

*Index Terms*— **Artificial intelligence, Machine learning, Biomedical imaging, Microscopy, Optics, Quantitative phase imaging**

## I. Introduction

THE last decade has witnessed the dramatic revival of two classic research fields in optics and computing, namely quantitative phase imaging (QPI) and artificial intelligence (AI). These seemingly unrelated disciplines have been studied separately for over half a century, but they share an intriguing historical feature. While the major concepts and ideas were proposed decades ago, their practical realization and commercialization became possible only recently owing to significant developments in the field of electronics, such as digital image sensors and graphics processing units (GPUs). These advances rapidly attracted the attention of researchers to the unexplored interface between the two fields, which has become extremely exciting for biomedical applications.

QPI deals with the "phase problem," which is a fundamental problem concerning the loss of phase information in physical measurements within the context of optical imaging [1, 2]. The high temporal frequency of visible light and the limited bandwidth of existing light sensors make a direct recording of optical phase information infeasible; typically, one can only measure the time-averaged signal, which is called intensity. Intensity-based optical imaging systems, including photographic cameras and even human eyes, are sufficient for daily living in the macroscopic world. However, the phase problem becomes crucial when one attempts to observe life in the microscopic world under a light microscope, which remains the tool of choice for live cell imaging. As most biological cells are transparent, and thus present minimal light absorption, a purely intensity-based imaging method, which is called bright-field microscopy, suffers from low contrast. To overcome this difficulty, researchers have devised various exogenous labeling strategies, including classic staining and recent genetic fluorescent tagging, which are time-consuming and may interfere with endogenous biological processes.

The high-contrast optical imaging of living cells without labeling was first enabled by Zernike's invention of phase-contrast microscopy in the early 1930s [3]. The technique converts the phase shifts induced by the higher refractive index (RI) inside cells, which cannot be readily visualized, into measurable intensity variations by manipulating the interference of scattered and unscattered light fields. Despite the success of phase-contrast microscopy and its variants such as differential interference contrast microscopy, conventional phase-imaging techniques cannot measure *quantitative* phase-delay maps owing to the cumbersome phase-intensity relations in these methods. In the mid-2000s, quantitative phase-microscopy techniques were proposed owing to the routine utilization of digital image sensors, including a charge-coupled device and a scientific complementary metal–oxide–semiconductor image sensor for biological microscopy [4, 5].

Manuscript received XX May 2018; revised XXX; accepted XXX. Date of publication XXX; date of current version XXX. This work was supported by KAIST, Tomocube, Inc., BK21+ program, and National Research Foundation of Korea (2015R1A3A2066550, 2017M3C1A3013923, 2014K1A3A1A09063027). YoungJu Jo and Hyungjoo Cho contributed equally to this work.

YoungJu Jo, Sang Yun Lee, Geon Kim, and YongKeun Park are with Department of Physics, Korea Advanced Institute of Science and Technology, Daejeon 34141, Republic of Korea (e-mail: astralatom@kaist.ac.kr; prism@kaist.ac.kr; gyugkei@kaist.ac.kr; yk.park@kaist.ac.kr). Hyungjoo Cho is with Graduate School of Convergence Science and Technology, Seoul National University, Suwon-si, Gyeonggi-do 16229, Republic of Korea (e-mail: phelahab@gmail.com). Gunho Choi and Hyun-seok Min are with Tomocube, Inc., Daejeon 34109, Republic of Korea (e-mail: ghc0311@gmail.com; min6284@gmail.com), along with YoungJu Jo, Hyungjoo Cho, and YongKeun Park.

Digitally recorded microscope images that are processed by computational phase retrieval techniques allowed access to quantitative phase images of living cells in two dimensions [6, 7]. Subsequently, robust two-dimensional (2D) QPI enabled the three-dimensional (3D) and even four-dimensional (4D, or time-lapse 3D) mapping of RI distribution by solving the RI–thickness coupling in phase images [8-12]. The fast and label-free nature of QPI presents a variety of advantages for biomedical applications, which are discussed later. The principles and recent progress of QPI are further reviewed in Section II.

AI is a broadly defined term that is utilized to designate artificial systems that mimic biological intelligence such as learning and problem solving by animals. Recently, AI research has been dominated by its subfield, called machine learning, which builds computer systems with data-driven learning ability [13]. Instead of being *explicitly* programmed or *rule-based*, a machine learning algorithm is designed to fit or optimize adjustable parameters in a computational model that reflects certain patterns in data through a learning rule. This approach is particularly useful when handling large-scale and high-dimensional data that hampers human investigation. Depending on the target tasks, machine learning is divided into three categories: supervised, unsupervised, and reinforcement learning, which are sequentially reviewed throughout this Review.

Machine learning has undergone recent advances owing to the renewed attention on artificial neural networks, which emulate biological neural networks in the brain [14]. While it has been known for decades that harnessing multi-layered neural networks as computational models is mathematically advantageous (as explained in detail in Section III), some fundamental difficulties with respect to training multiple layers and the large number of learnable parameters have made them infeasible for practical applications [15]. Since the mid-2000s, powerful computing resources based on general-purpose GPUs, large-scale datasets such as ImageNet, and clever training algorithms by Hinton among others have together revolutionized the field, which is now called *deep learning* [14, 16, 17]. The remarkable learning ability of deep neural networks, which significantly outperforms conventional machine learning techniques, has been demonstrated in a variety of disciplines, including computer vision.

When QPI meets machine learning, there are unexpected advantages that arise owing to the inherent characteristics of QPI. In conventional labeling-based imaging techniques such as fluorescence microscopy, the data characteristics are highly dependent on sample preparation protocols. In contrast, QPI relies on endogenous RI contrast, which is invariant under experimenter or instrument variations [18]. This high-uniformity data can be obtained even at scale because QPI is fast and label-free. In short, QPI inherently generates uniform-quality and large-scale data, which are ideal for exploring with machine learning (see Section III). Furthermore, machine learning can enhance QPI itself by learning aspects of the underlying physics (see Section IV). In this Review, we explain the synergy between QPI and machine learning, review recent and rapidly expanding literature, and provide practical guidelines and perspectives for further development.

## II. QUANTITATIVE PHASE IMAGING

QPI is a class of light microscopy techniques that enable the visualization of quantitative optical field maps, i.e., both amplitude and phase information. While merely imaging light intensities, as is the case with typical optical cameras, is straightforward (Fig. 1(A)), the retrieval of phase maps determined by RI distribution requires special methods. Among the various techniques for 2D phase retrieval, in this Review, we focus on holographic QPI based on spatially modulated interferometry in transmission geometry [19], but the rest of this Review applies to all techniques. Technical details regarding different methods, such as the transport of intensity equation, ptychography, phase-shifting interferometry, in-line holography, and reflection-geometry QPI, can be found elsewhere [2].

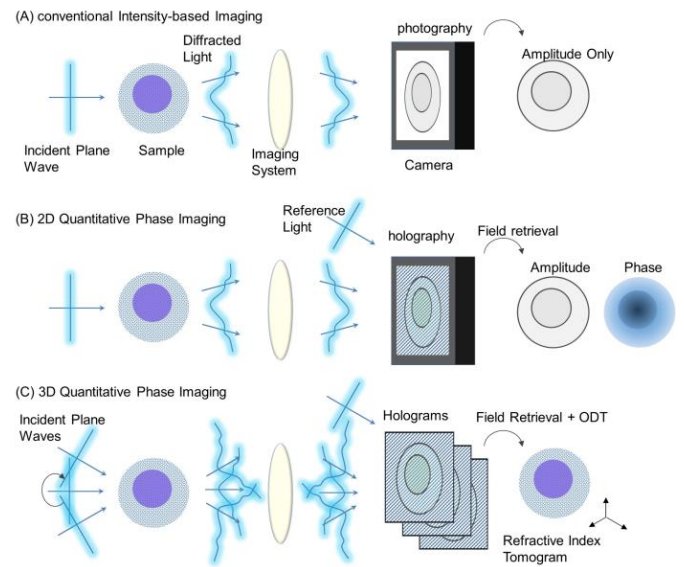

Fig. 1. Concepts of (A) conventional intensity, (B) 2D quantitative phase, and (C) 3D quantitative phase imaging techniques.

### A. Two-dimensional Quantitative Phase Imaging

A typical setup for 2D QPI is presented in Fig. 1(B). To measure the optical wavefront of the light diffracted by the sample, we record the interference pattern using a reference light with a known wavefront. In this way, usually undetectable phase information is converted into a detectable fringe pattern, which is called a hologram. Then, both the amplitude and phase of the measured light can be retrieved from the hologram by using computational field retrieval algorithms [6, 7]. If the sample is a phase object such as a single cell, the retrieved field information is a phase-delay map induced by the RI difference between the object and the surrounding medium. The RI distribution and the induced phase map carry both structural and chemical information of the object [18].

As briefly summarized in Section I, most microscopic biological cells are transparent in the optically visible wavelength range, and thus do not present adequate contrast for conventional bright-field imaging. Therefore, QPI is particularly advantageous for the label-free imaging of live cells, and has thus been widely utilized for biomedical applications [1, 2, 18]. Meanwhile, QPI has been criticized because of its limited chemical specificity; resolving different molecular species with similar optical properties is not readily apparent. To overcome this limitation, in Section III, we focus on machine learning approaches.

*B. Three-dimensional Quantitative Phase Imaging*

QPI is even more powerful for the 3D imaging of transparent microscopic objects. The optical field maps measured with varying incident angles, as shown in Fig. 1(C), can be utilized for the reconstruction of the RI distribution in three dimensions by employing optical diffraction tomography (ODT), which was formulated by Wolf in the late 1960s [8, 11, 12, 20]. The governing equation for this inverse problem is the Helmholtz equation, which describes light propagation in matter. ODT retrieves the 3D RI distribution that satisfies the governing equation using the measured optical fields. Specifically, a set of measured 2D optical fields with various illumination angles is successively mapped in a 3D Fourier space for the scattering potential of the sample, which is equivalent to the corresponding RI distribution. Note that here we assume weak scattering - slightly varying permittivity at a wavelength scale. Inaccessible information due to finite aperture sizes, also known as 'missing cone problem', is typically estimated using regularization-based iterative reconstruction techniques [21]. Then, the inverse Fourier transform gives the 3D RI distribution, i.e., the RI tomogram.

Simple analogies for the 2D and 3D QPI techniques are 2D X-ray imaging and X-ray computed tomography (CT), respectively, both of which are performed in hospitals on a daily basis. In X-ray CT, it is assumed that X-rays follow straight paths in space. However, to precisely reconstruct the 3D RI tomogram of a sample in the visible light range, light diffraction should be properly accounted for, as in ODT.

The representative 2D and 3D QPI images are shown in Fig. 2. A commercial ODT system (HT-2H, Tomocube Inc., Republic of Korea) was used. The setup is based on off-axis Mach-Zehnder interferometry equipped with a DMD [22]. The sample is live human adenocarcinoma cells (MDA-MB-231, ATCC, United States). Cells in a buffer medium are sandwiched between two coverslips before the measurements. No fixation or other preparations was used.

In ODT with coherent light, the measurement of 2D optical fields is conducted either by varying beam-illumination angles using galvanometer mirrors [23, 24] or spatial light modulators (e.g., digital micromirror device [22]), by rotating the sample [9, 25], or by scanning the sample with different wavelengths [26, 27]. In addition, 3D RI tomography is available with partially coherent [28] or incoherent [29] light that has less coherent or speckle noises, in the cost of additional sample-scanning processes in the axial direction owing to its low coherence length. Fast angular scanning and GPU-based reconstruction techniques enable even time-lapse 3D ODT, or 4D QPI [30].

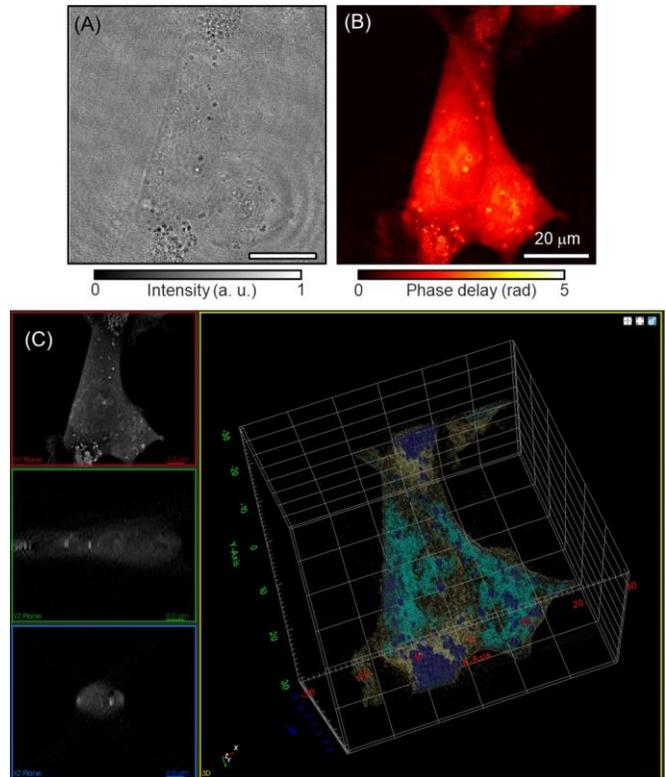

Fig. 2. Representative QPI images of biological cells. (A-B) 2D QPI images. The (A) amplitude and (B) phase image of the cell. (C) 3D QPI image of the cell. (Left column) The cross-sectional images of the 3D RI tomogram. (Right column) the 3D rendering of the RI tomogram.

Recently, ODT has been applied to various applications, including biophysics [31], hematology [32], microbiology [33], infectious disease [34], neuroscience [35], cytotoxivity [36], and biotechnology [37].

III. USE OF MACHINE LEARNING TO HARNESS AND UNDERSTAND QPI DATA

Recent advances in machine learning have opened an exciting and rapidly growing frontier with respect to the systematic and thorough utilization of QPI data. First we clarify the advantages of using machine learning to handle QPI data. A major benefit is from the nature of biomedical QPI: *fast acquisition in the cost of low chemical specificity*. In essence, QPI is a label-free imaging modality that exploits the endogenous distribution of RI over space and time as its imaging contrast [1, 18]. Because the RI distribution is governed by the structural and chemical properties of the samples, QPI enables quantitative and simultaneous measurements of the relevant biophysical properties in various cellular and subcellular systems. For instance, morphology (volume and surface area), dynamics (membrane fluctuation and motility), and biochemistry (biomolecular density and dry mass) can be quantitatively addressed at the individual cell level.

While QPI is an unparalleled method to measure the morphological and dynamical features, in particular in 3D and 4D, the chemical specificity of the endogenous RI-based imaging is limited compared to conventional direct labeling-based techniques such as fluorescence microscopy. This limitation primarily comes from the difficulty with mapping the RI distribution to the chemical composition. To date, the most successfully obtainable chemical information from QPI is the dry mass, whose pointwise proportionality relative to the RI value was well-characterized more than half a century ago [38, 39]. This mass-RI relation's coarse-grained character, i.e., minimal sensitivity to the constituent chemical identities, implies that inferring the chemical composition requires full consideration of the RI *distribution* or *spatial contexts*, rather than pointwise values or gradients [22]. This type of pattern-recognition problem is central to the field of machine learning. Machine learning is a powerful tool to augment the chemical specificity of QPI in a data-driven manner powered by high-throughput and uniform-quality data acquisition through label-free measurement.

*A. Classification: Conventional Approaches*

The benefits of machine learning have been demonstrated in a major QPI application: the classification and identification of cells and tissues for rapid screening and diagnostic purposes. Classification in computer vision is a problem to build and train a *classifier* that maps an input image to the corresponding sample classes (e.g., species or cell types, in a biomedical QPI context). Solving a classification problem typically comprises two stages: training and testing. During training, the images (or extracted features) with the corresponding ground truth class labels (training data) are utilized to train, or optimize, the parameters in a classifier in a supervised manner (i.e., supervised learning). An optimization procedure typically accompanies a carefully designed loss function and the corresponding learning rule.

As explained above, merely considering one or two simple features in QPI is often not sufficient for the high-specificity discrimination of the samples, mainly because of the intrinsic characteristics of QPI and the heterogeneity in biological systems. The training procedure enables the systematic integration of various characteristics encoded in the RI distribution to extract the class-dependent fingerprints over intra-class variations. In this way, it is possible to learn the subtle but present information regarding distinguishable chemical identities, and thus to augment chemical specificity computationally.

Although training may be time consuming, depending on the training data size and classification model complexity, after training, the fast and label-free nature of QPI allows rapid identification. The learned fingerprints are harnessed to automatically identify, or make class predictions for, the newly measured images (test data) upon deployment of the classifier. The performance can be quantified by comparing the predicted and ground truth labels. When the target classes for classification share highly similar chemical, morphological, or genetic traits, the classes appear to be indistinguishable for human investigation, and thus the classification schemes for QPI are often designed for super-human performance. This classification approach is also applicable to relatively easier problems that require human-level performance for automated ultra-high-throughput interrogation.

The history of classification techniques for QPI data dates back to the mid-2010s, when biomedical QPI was at its initial stage. In their paper published in 2005, Javidi *et al.* reported the QPI-based recognition of two algae species [40]. First, two-dimensional optical field images were numerically reconstructed at multiple depths. From these semi-3D data, they extracted Gabor wavelet-based features that represent both spatial frequencies and local properties, and then trained the species classifiers based on rigid graph matching. After this pioneering paper, they extended their work using improved shape-tolerant feature-extraction methods [41, 42], and to other applications including cyanobacteria [43] and stem cells [44]. A comprehensive review summarizes their series of papers [45]. After a decade, this topic was revived, and started to attract the interest of researchers, mainly because of the significant advances in machine learning. In recent years, a variety of samples in biomedical contexts were interrogated by QPI combined with classification techniques: pathogenic bacteria [46, 47], lipid-containing algae [48], beads versus cells [49], abnormal or infected red blood cells (RBCs) [50-53], yeast [54], lymphocytes [48, 55], macrophages [56], sperm cells [57], melanoma cells [58, 59], breast cancer cells [60], circulating tumor cells [48], cancer tissues [61-66], and even air-borne particulate matter [67].

Whereas all these studies share the basic classification framework, some of them present notable variations in terms of input data types and output classes, and thus open up a path to exciting new applications. Three-dimensional RI tomograms, which directly uncouples the RI–thickness coupling in conventional 2D QPI using multi-angle or multi-plane measurements, were used for the label-free sorting of lymphocytes [55]. The systematic exploration of the abundant information in 3D tomograms has just begun. *Time-lapse* observation exploiting the label-free and minimally phototoxic nature of QPI was utilized to monitor the kinetic state transitions in melanoma and breast cancer cells, establishing tools to study the epithelial-mesenchymal transition, which is central to metastasis in cancer progression [58-60]. These studies, along with others, demonstrate that dynamic cellular states, including kinetics, activation, and viability are manageable and fruitful targets for QPI-based classification schemes [54, 56, 58-60]. *Spectroscopic* QPI, which was employed to diagnose malaria [51], may provide a new dimension of information to be explored further, probably along with *polarization-dependent* information [68, 69]. *Scattering* information, which is inherently obtained in QPI via Fourier transform light scattering [70], was used for the rapid identification of bacterial species to cope with acute infection [46, 71]. This study was inspired by and provided design insights to angle-resolved light scattering measurements in bulk or flow cytometry settings [72, 73]. For *high-throughput* QPI measurement that is crucial for the data-driven approach, microfluidic platforms and the photonic time-stretch system were integrated [48, 74, 75]. For *point-of-care* applications, miniaturized and on-chip QPI devices were developed in a compact, portable, and cost-effective fashion [47, 54, 67, 76-

78]. Finally, a recent report of QPI-based *augmented reality* applications also provides new insights into future applications [79].

### B. Classification: Deep Learning Approaches

Despite the increasing number of classification-based studies in QPI, most of these have relied on problem-specific design and extensive domain knowledge by using *hand-designed feature extraction*. For example, the measurement of the spatiotemporal fluctuation of RBC membranes, which has been important in soft matter physics and disease-diagnosis applications, is based on the homogeneity of the intracellular hemoglobin concentration [80]. In addition, the tracking and quantification of lipid droplets in eukaryotic cells utilize the remarkably high RI of lipid [81]. In short, a simplified biophysical model designed by the domain experts is required for each biological system of interest to enable efficient feature extraction. For complex systems or measurements that preclude simple modeling (e.g., 3D dynamics of eukaryotic cells), one typically relies on cumbersome trial-and-error processes to combine the basic features agnostic to the system characteristics [55]. This fundamental limitation has hindered the ability of conventional classification schemes to address a wide range of biological systems and biomedical applications, especially for non-experts.

As introduced in Section I, the application of deep learning is currently transforming the field. While a single layer of neurons is equivalent to a linear classifier, layering them into a multi-layered network, with threshold-like nonlinear functions between the layers, makes the network mathematically flexible [15]. Deep neural networks have been proven to be capable of approximating virtually any arbitrary function if one properly allocates and adjusts the synaptic weight parameters between the computer-simulated neurons (universal approximation theorem). The remarkable flexibility of deep neural networks combined with recent advances in learning algorithms, GPU computing, and large-scale datasets enables significant learning abilities, outperforming the conventional machine learning methods that rely on linear or slightly nonlinear basis functions in a variety of disciplines [14].

Importantly, deep neural networks can be directly trained using raw data without manual feature extraction. As a canonical example, the most successful and widely used deep-learning architecture, namely the convolutional neural network (CNN), learns hierarchical representation reminiscent of visual processing in the retina and visual cortex [17]. Progressing through the layers in the network, data dimensionality decreases, while the degree of abstraction increases, or vice versa. In convolutional layers of a CNN, each neuron acts as a *learned feature detector* via the convolutional filtering of input data with the synaptic weights that possess the localized and shared receptive field (or spatial support). Because these feature detectors are learned but not hand-designed, deep neural networks are considered to have the unique capability of feature learning (or *representation learning*). This advantage eliminates the need for hand-designed feature extraction, and thus facilitates a genuinely data-driven approach through end-to-end learning.

The first application of CNN in QPI was recently demonstrated in a classification scheme for the rapid optical screening of anthrax spores [47]. As illustrated in Fig. 3, a specialized CNN named HoloConvNet was designed and trained to discriminate *Bacillus anthracis* spores from other *Bacillus* species with high genetic similarity. To train the deep neural network, a variety of classic and recent techniques were synergistically combined, e.g., batch normalization, dropout, rectified linear unit (ReLU), momentum, data augmentation, error backpropagation, and grid-based hyperparameter search [15, 82-84]. Despite the seemingly indistinguishable phase images, HoloConvNet identified anthrax spores with high sensitivity and specificity. Subsequent control experiments showed that the network automatically recognizes and exploits inter-species dry mass differences that may arise from subtle structural characteristics. It is interesting that HoloConvNet was not explicitly taught to calculate dry mass from phase images; the remarkable representation learning capability enables automatic feature extraction to boost the chemical specificity of QPI. This advantage is further validated by the successful utilization of the identical network and hyperparameters for a different application: optical diagnosis of the pathogen *Listeria monocytogenes*, whose dry mass is not a key discriminating feature. In short, deep learning opens up a route to QPI applications in complex biological systems via modeling-free investigations, even by non-experts.

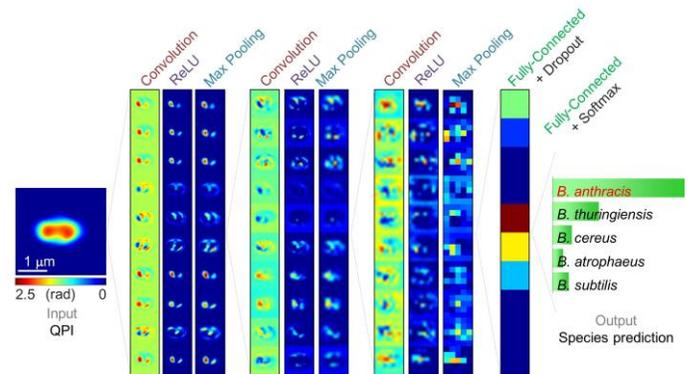

Fig. 3. Architecture of HoloConvNet, the first CNN-based classification network for QPI. Reproduced with permission from [47].

### C. Segmentation

In computer vision, it is crucial to identify the pixels belonging to particular regions of interest (see Fig. 4). In contrast to classification, which is image-wise recognition (image-to-class), *segmentation* is pixel-wise recognition, which estimates class probabilities for each pixel (image-to-image). Segmentation is mandatory for the quantitative analysis of QPI data [80, 81]. Conventional hand-designed segmentation algorithms for QPI have been mostly based on thresholding by pointwise values or gradients.

Machine learning provides new opportunities for this topic. A recent paper reports the QPI-based automatic segmentation of prostate cancer tissues to regions with different severity levels [61]. As demonstrated in this study, machine learning can be particularly powerful for QPI-based digital histopathology, which precludes conventional segmentation techniques. At the cellular level, the more accurate segmentation of RBCs was achieved using a CNN variant called the fully convolutional network [85]. In addition, it may be clinically useful to extend this technique into multiclass segmentation when investigating heterogeneous cell populations such as blood.

A series of recent papers from related disciplines indicate the potential for even more exciting applications at the subcellular level. By learning the mapping from conventional (e.g., bright-field and phase-contrast microscopy) to fluorescence microscopy images, the label-free visualization and segmentation of intracellular organelles were demonstrated [86-88]. In this trans-modal approach, fluorescent markers were utilized to generate the ground truth images (see Section IV from a regression point of view). Because QPI measures much more abundant optical information compared to the conventional microscopy techniques, it is strongly expected that recent correlative multimodal QPI approaches would exhibit superior performance [89-91]. In this way, one may explicitly demonstrate that QPI has a significant chemical specificity that enables various new applications.

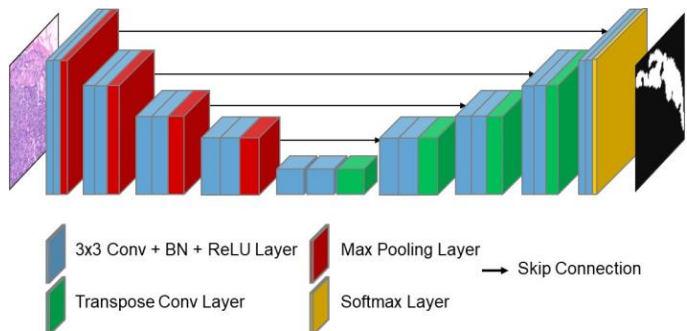

Fig. 4. Schematic for CNN-based segmentation network for QPI. Note the presence of transpose convolution layers in contrast to Fig. 3.

### D. Unsupervised Learning

Whereas the aforementioned machine learning approaches mostly feed the ground truth (e.g., class labels for classification, object location for segmentation) to the algorithms (supervised learning), learning without the ground truth also constitutes an important class of machine learning (unsupervised learning). The absence of the ground truth fundamentally distinguishes the nature of unsupervised learning from the supervised counterpart; rather than learning the input-output relation, unsupervised learning methods typically optimize loss functions that are designed for specific uses.

A crucial application is dimensionality reduction for data visualization or compression. A traditional algorithm that satisfies both of these goals is principal component analysis (PCA), which is equivalent to an autoencoder when its decoder is linear and the loss function is the mean squared error (MSE). The aforementioned HoloConvNet paper includes several plots that use t-distributed stochastic neighbor embedding (t-SNE), which is a popular high-dimensional data visualization technique for deep learning [47]. Stacked autoencoders and restricted Boltzmann machines, which were originally proposed for data compression, have significantly contributed to the revival of neural networks via unsupervised layer-wise pre-training for the efficient supervised training of deep networks. Another application is exploratory data analysis using clustering techniques for grouping the data points with similar properties.

In contrast to supervised learning, neural network-based unsupervised learning is a largely unexplored area, and may provide new opportunities upon maturity.

### E. New Data and Methods

Recent advances in QPI techniques provide exciting new data to be explored with machine learning. Above all, 3D tomographic data would be the most straightforward extension. Tomographic cell-type classifiers that are based on conventional machine learning techniques were reported within the context of label-free lymphocyte sorting, providing a testbed for the deep-learning approach in 3D [55]. While 3D CNNs that employ 3D convolutions have been proposed in other disciplines [92, 93], their applications to 3D QPI remain unexplored. Another obvious application is with respect to time-lapse data. As QPI is free from photobleaching and phototoxicity, time-lapse imaging at various time scales from seconds to days is available [30, 94, 95]. Instead of frame-wise analysis, employing recurrent neural networks (RNNs) such as long short-term memory (LSTM) and gated recurrent unit (GRU) would properly address the inter-frame relations [96, 97]. A combination of CNNs and RNNs can be particularly useful to deal with the 4D QPI data. It is desirable to establish a universal deep-learning framework for the principled handling of various QPI data by considering the fundamental characteristics of RI.

A common criticism related to deep learning is that a deep neural network is a "black box" that is not interpretable. However, recent techniques for visualizing and understanding the inner working of deep networks enable the interpretation of the high-performing networks [98-100]. Implementing machine attention also facilitates systematic interpretation [101, 102]. This type of approach may assist researchers to discover interesting new patterns or testable hypotheses from large-scale QPI data.

## IV. USE OF MACHINE LEARNING TO ENHANCE QPI

Machine learning also enhances QPI itself, mostly in terms of efficiency and performance. In essence, all measurements rely on respective theoretical models based on the physical principles and specific assumptions. For example, as explained in Section II, ODT is an elaborate framework for inverse scattering based on Helmholtz equation and weak scattering assumption. Obviously, such hand-crafted models are straightforward but require significant domain knowledge and are limited by the assumptions. Alternatively, now one can replace modeling by machine learning some aspects of the underlying physics in a data-driven manner.

While in the preceding section we focused on classification and segmentation, in this section, the primary technique is image-to-image *regression* that predicts pixel-wise values. Recent advances in deep learning and brilliant new ideas are accelerating progress in this direction.

### A. Phase Retrieval and Tomographic Reconstruction

As described in Section I, computational phase retrieval algorithms are core techniques for QPI and related disciplines that deal with the phase problem [103]. While conventional algorithms are explicitly based on the underlying optical principles [6], data-driven phase retrieval techniques learn the correspondence between the measured intensity-phase pairs that were recently developed for in-line holography [104, 105] and Fourier ptychography [106-108]. These algorithms are often significantly faster than conventional ones by removing the time-consuming Fourier transform or iterations. Moreover, certain unprecedented features, such as single-plane phase recovery for on-chip holography, are also available..

Machine learning-based tomographic reconstruction algorithms are also being rapidly developed as in phase retrieval. The conventional framework consisting of phase retrieval, ODT, and iterative reconstruction is time-consuming, even with GPU acceleration [10, 21, 30]. As in X-ray CT and other related inverse imaging problems [109, 110], CNN-based 3D RI tomography recently demonstrated faster reconstruction via end-to-end processing [111]. We anticipate that the next step is to utilize the training data prepared by conventional reconstruction algorithms that are slower but more accurate than ODT [112, 113]. This approach may enable high-speed 3D RI tomography that directly addresses multiple scattering, as in thick tissues.

Although these approaches learn the underlying physics to some extent and present compelling advantages, one should be careful when dealing with new types of samples. Both phase retrieval and tomographic reconstruction are often fundamentally ill-posed inverse problems lacking uniqueness of the solutions (e.g., single-plane phase recovery in on-chip holography, missing cone problem in ODT). In such cases, the algorithms would learn specific characteristics of the training data and may show poor performance for new data.

### B. Image Enhancement

Realizing the computational enhancement of QPI images has been a long-standing endeavor since the birth of digital image acquisition. As in phase retrieval and tomographic reconstruction, the development of machine learning-based image enhancement has opened a new frontier.

An important feature of QPI is computational refocusing through the numerical propagation of optical fields. Because this calculation is also time-consuming, researchers have devised fast autofocusing using machine learning [114-116]. A related technique is the learning-based holographic tracking and characterization of particles [117-119].

Also, suppressing coherent or speckle noise inherent in most QPI images may benefit from the data-driven approach. While the conventional techniques relied on measuring multiple images with angular, spectral, and polarization diversity [120] or hand-crafted denoising algorithms [121], learning-based methods are expected to provide a more efficient methodology.

There have been more rapid advances in image-enhancement methods that are relevant to both QPI and general microscopy techniques, including aberration correction [122-125] and depth-of-field extension [115, 123, 126]. The introduction of CNN-based resolution improvement [126, 127] or generative adversarial network (GAN)-based style transfer [128, 129] to QPI may be intriguing, but should be carefully addressed.

### C. Design and Control of Imaging Systems

All of the literature reviewed so far focused on post-measurement analysis. A recent study demonstrated that machine learning can facilitate the design of optical imaging systems as well [107]. Training a CNN-based classifier, with a learnable layer simulating image formation in Fourier ptychography, provided an effective design for light source configuration. This machine-driven design approach simultaneously optimizes both the imaging setup and the post-processing algorithm (e.g., classification) in a synergetic manner. Extending this strategy to other QPI techniques would provide fruitful advances.

Recent approaches in related disciplines indicate that machine learning can also improve QPI during measurement through reinforcement learning, which is another class of machine learning [130-132]. Reinforcement learning is a machine learning problem that is used to train the machine to take actions in environments to maximize the reward [133]. As the space of measurement parameters (e.g., the angle of illumination in ODT) is often huge and redundant, the on-line control of a QPI setup using deep reinforcement learning may result in an efficient and ultrafast measurement. The on-chip implementation of deep neural networks would facilitate this type of closed-loop experiment [134].

## V. PRACTICAL GUIDELINES FOR DEEP LEARNING IN QPI

Deep learning is an important technique that is employed to explore biomedical images from a variety of modalities, including magnetic resonance imaging (MRI), X-ray CT, electron microscopy, and light microscopy [135-139]. Compared with other imaging methods, deep learning applications in QPI have been less explored. To guide practical applications in this direction, here, we describe a typical pipeline that considers the general characteristics of biomedical images as well as QPI-specific perspectives.

### A. Problem Definition

First, one should decide the problem to be solved. The most common deep learning tasks in biomedical image analysis are classification and segmentation [137, 140, 141]. While both have been described in the QPI context in Section III, here, we summarize their recent trends for biomedical imaging in general.

Classification: The high-performing architectures proposed in other domains have been introduced to biomedical image analysis. Currently, the most widely used models are ResNet [142] and Inception-v3 [143], which implement efficient deep structures. Both models proposed tricks for learning deep networks: short-cut connection with a bottleneck and auxiliary loss with inception module, respectively. Subsequently, more powerful architectures, such as DenseNet [144], which added pre-activation and dense connectivity, and ResNeXt [145], which added group convolution by combining short-cut connection and inception, have been proposed. Architectures that reflect all of these features have also been proposed, and further developments are being made rapidly. The most typical loss function for classification is conventional cross entropy. For multi-label classification, ranking loss can be a good choice [146].

Segmentation: The baseline model U-net exhibits good performance for segmentation [147]. To enhance its output quality, the following approaches have been devised. First, increasing the depth of networks using short-cut connection or dense connectivity is a method that is employed to minimize information loss and to enhance the output quality [144]. Alternatively, the use of dilated convolution prevents information loss by downsizing and considers receptive fields with various sizes [148]. Recently, a combination of both methods was also proposed; a dilated convolution learns a small feature map considering receptive fields with various sizes, and then a decoder effectively up-scales and connects it with an encoder and skip connection [149]. Typical loss functions are cross entropy and its variants, as in classification. For binary segmentation, image similarity metrics such as the Dice similarity coefficient and Jaccard index are also utilized. In the case of multiple classes, the mean intersection over union (mIOU) is typically used.

### B. Data Preparation

In general, the data characteristics of biomedical images are dependent on the variations in sample preparation, instruments, and experimenters. Because these alterations typically cause performance degradation in machine learning on the data, preprocessing techniques such as registration, normalization, and image enhancement are necessary [150-153]. Registration and normalization align different data into the same coordinate and range, respectively. Image enhancement reduces noise and improves image quality for more accurate analysis. There are two representative approaches for these procedures: optimizing a predefined metric [154, 155] and generating transform parameters or transformed images using separate networks [156-158]. As mentioned in Section I, QPI that is based on endogenous RI distribution is mostly free from the cumbersome registration or normalization. However, multimodal or correlative QPI approaches may require preprocessing for non-QPI channel data [90].

It is also important to prepare the data appropriately in terms of efficiency and performance. Because medical images are typically large in size, the limitations in GPU memory may incur penalties with respect to network capacity and mini-batch size. For image-to-image inference tasks such as segmentation, the patch-based approach can be used to overcome this difficulty. Dividing a single image into multiple patches enlarges the training set regarding the number of images while addressing the memory concern. However, the patch-based approach may degrade performance as it considers local features only. In the case of classification tasks, the whole image-based approach is preferred when considering both local and global features as well as reducing the inference time [159]. Recently, it was proposed to diversify the patch size or hybridize the two approaches using a cascade structure [157]. Another concern relates to how to treat the data in 3D, 4D, or beyond. For instance, 3D data can be prepared either as stacked 2D image channels or as being genuinely volumetric [92, 93, 136, 147, 160, 161], and this choice should be consistent with the model design.

### C. Model Design

A neural network model comprises a neuron model, network architecture, and learning rule. It is essential to carefully design the architecture to match the data complexity and to avoid overfitting or underfitting. The learning rule accompanied by a set of hyperparameters is also crucial for proper training. Here, we describe a set of key design considerations for a deep learning model.

Activation function: The primary role of the activation function, which is a key property of a neuron model, is to introduce non-linearity. In addition to the most representative activation function called ReLU [84], there are also its variants, including leaky ReLU [162], parametric ReLU [163], and absolute value rectification. The most generalized form of ReLU is Maxout [164]. RNNs such as LSTM or GRU may still use conventional sigmoidal functions [96, 97]. They resolved the vanishing gradient problem, which is a chronic issue for sigmoids, through innovative architectures. Further, there is softplus [165] and the exponential linear unit (ELU) [166], which are approximate differentiable forms for ReLU and leaky ReLU, respectively. Because the performance of an activation function is dependent on the problem setting, it is necessary to select an optimal activation function by performing comparative experiments. However, ReLU exhibits stable performance in most cases.

Kernel size: In CNNs, the kernel size indicates the size of the receptive field for a neuron [167]. The majority of recent architectures employ the size of 3-by-3 for weight factorization; larger kernels can be replaced by a series of 3-by-3 kernel operations, reducing the memory requirements. Likewise, a 3-by-3 kernel can be decomposed into 1-by-3 and 3-by-1 kernels [143].

Pointwise convolution or bottleneck: When the depth of feature maps increases as it progresses through the layers, pointwise convolution or bottleneck may boost the learning efficiency and computational speed by reducing the number of required parameters [142].

Short-cut connection: It is possible to train deeper networks with high performance using short-cut connections. A simple but powerful implementation is identity mapping, which adds an input to the output of a layer [142]. In an encoder-decoder network, the skip connection using inter-layer identity mapping at the same depth can be utilized [147, 168]. Recently, it was also proposed to use concatenation instead of identity mapping [144].

Loss function: In deep learning, the loss functions are essentially similar to those used in conventional parametric models. The most representative examples are the MSE and cross entropy, assuming Gaussian and multinoulli distributions, respectively. Depending on the context, the mean absolute error (MAE) or negative log-likelihood (NLL) are also employed. Importantly, the choice of a loss function determines appropriate network output units, and should therefore be task-dependent. For a classification task, cross entropy with softmax is relevant. If the classes are imbalanced, weighting methods, such as weighted cross entropy, focal loss, and Tversky loss, can be employed. For a regression task to predict a certain quantity, MSE or MAE together with linearity or ReLU can be used. For a regression task to predict the probability, NLL along with sigmoid is appropriate. For multimodal regression tasks, which have attracted interest recently, a mixture density network with context-dependent output units may be useful [169].

Regularization: To prevent overfitting, one can introduce additional assumptions for the optimization process. Typical strategies add new terms to the loss function with tuned coefficients controlling the regularization strength. A conventional but useful technique is parameter norm penalty, which is also called weight decay [15]. Lasso, ridge, and elastic regularization belong to this technique. In addition to the classic methods, regularization effects can be obtained by early stopping, ensemble learning (or bagging), noise injection (to input, output, or parameters), and noise robustness loss [170]. Recently, dropout [82], batch normalization [83], and shake-shake regularization using short-cut connections [171] have been proposed as well. Multi-task and multi-stage learning, which are based on parameter sharing and tying, respectively, are also effective [171, 172].

Optimizer: Training a neural network is essentially an optimization process that minimizes the loss function according to a learning rule. While conventional error backpropagation based on stochastic gradient descent (SGD) is still powerful, several variants, such as momentum-based methods, have been proposed [15]. The adaptive momentum estimation (Adam) optimizer, often followed by SGD, has been a practical and effective choice in recent years [173]. It is worth noting that a suitable weight initialization may enhance optimization [163].

Batch normalization: Unbiased input data distribution to each layer enables faster and more robust training. Batch normalization has resolved this so-called covariance shift problem through the layer-wise normalization of input data distribution by learning the scale and shift factors [83]. As previously mentioned, the technique also acts as a regularization method. Nowadays, batch normalization is a standard in most deep learning models.

Automatic search for optimal architectures and hyperparameters: Although deep learning has freed researchers from the need for manual feature design, it instead often requires a manual search for optimal architectures and hyperparameters. Several studies have attempted to automate these procedures. Most approaches rely on reinforcement learning or genetic programming for empirical search [174-176]. While these strategies require significant computational resources, a new method based on parameter sharing exhibits efficient automation [172]. However, universal automation techniques that work across domains are yet to be realized. Thus, it is still helpful to construct a model that reflects domain insights.

*D. Training and Evaluation*

The proper split of the dataset should precede the training and evaluation of a deep learning model. In supervised learning, the dataset is divided into three disjoint subsets. These are training, validation, and test sets (for simplicity, the explanation of the validation set was intentionally omitted in Section II). A training set is used to train the model. Then, the performance can be temporarily evaluated using the validation set to optimize the architecture and hyperparameters. After optimization, the test set is utilized to evaluate the final performance of the model. One should be careful to avoid a biased split of the dataset while securing a sufficient size for each subset and considering target applications [177].

The performance of a deep learning model depends largely on the amount of labeled high-quality data for training. Despite the importance of gathering sufficient data, it is often limited in many biomedical applications [178]. Several strategies have been devised to overcome this challenge. One can utilize transfer learning, which builds on a model pre-trained with non-biomedical images [142, 179]. In addition, data augmentation may be employed to enlarge the training set by performing computational transforms or added noise. While simple transforms, such as rotation, flipping, cropping, and resizing, are mostly used [180], advanced techniques such as affine transform and elastic deformation can also be useful, depending on the data characteristics [147, 181, 182]. Data augmentation may also be advantageous to relieve class imbalance. It is also possible to re-learn using hard samples whose machine predictions are wrong. This approach ensures maximum generalization from limited data [183-185]. In the case of limited annotations, unsupervised or semi-supervised learning can be utilized [186, 187].

In addition to the loss functions described above, various evaluation metrics may be used depending on the context and target applications. For binary classification, the receiver operating characteristic (ROC) and precision-recall curve have been widely used to address the sensitivity-specificity tradeoff. In this case, one can set an optimal classification threshold using the F1 score and estimate the performance using area under the curve (AUC). For multiclass classification, one can use confusion matrix, among many others. For regression, MSE and MAE, as well as many application-specific image metrics such as structured similarity (SSIM) and peak signal-to-noise ratio (PSNR), can be employed. In addition to the metrics, it is often useful to evaluate a deep learning model using human experts [131, 132, 177].

VI. OUTLOOK

We reviewed the exciting frontier at the interface between QPI and AI. Because the remarkable synergy at the interface is owing to the inherent characteristics of QPI, this rapidly growing field is expected to provide an indispensable toolbox for QPI. Further, recent commercial QPI systems will even accelerate the development by making QPI more accessible to biomedical experts. Collaboration between the experts from QPI, machine learning, and biomedicine will result in novel applications.

Establishing publicly available standard datasets, such as ImageNet in computer vision, would significantly facilitate further developments in the field. Currently, published algorithms are mostly based on experimental data generated in individual research groups. Standard datasets would enable comparing different algorithms and thus guide new investigations.

It is also important to recognize the limitations and potential pitfalls of the data-driven approach. While machine learning is a powerful tool for discovering useful patterns in the data, it is impossible to find a pattern that does not exist (see the discussion on ill-posed inverse problems in Section IV). One should carefully consider whether or not a target application benefits from the data-driven approach.

From the physical point of view, the AI-aided QPI approaches can also be expanded to the field of wavefront shaping techniques [188, 189], because many of wavefront shaping techniques deal with the measurement and modulation of optical field information. For example, imaging through turbidity has been demonstrated by employing convolutional neural network [190, 191], and demonstrated several advantages over conventional approaches based on the measurements of light transport information [192, 193].

Moving forward, we envision that the synergistic combination between QPI and AI could have far-reaching applications in biomedicine, potentially in combination with newly emerging image-based cell profiling [194], rapid imaging cytometry [195], and correlative imaging [90, 196].

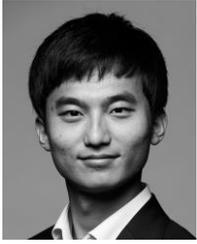
**YoungJu Jo** is an undergraduate student in Physics and Mathematical Sciences at KAIST, South Korea. He will start his doctoral studies in Applied Physics at Stanford University in 2018 Fall. He has been working in Prof. YongKeun Park's research group since 2012. In particular, he has led the group's machine learning approach to biomedical quantitative phase imaging. He also works for the start-up company Tomocube. He has been supported by Asan Foundation Biomedical Science Scholarship (2018), SPIE Optics and Photonics Education Scholarship (2014), and KAIST Presidential Fellowship (2013). He is a winner of KAIST Creativity and Challenge Award (2018), Samsung HumanTech Paper Award (2017), and Talent Award of Korea (2015).

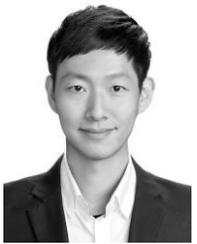
**Hyungjoo Cho** is a researcher at Tomocube. He received the B.S. degree in Electronic Engineering from Dongguk University, South Korea, and currently is a Master's candidate in Graduate School of Convergence Science and Technology, Seoul National University, South Korea. Previously, he was a software engineer with LG Electronics, where he was in charge of developing gesture recognition algorithm using optical devices. His research interests include machine learning approaches for solving biomedical imaging problems.

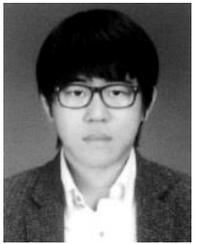
**SangYun Lee** was born in Seoul, South Korea, in 1990. He received the B.S. degree in Physics from KAIST, South Korea, in 2013. Since 2013, he has been studying in the Ph.D. program in the Department of Physics at KAIST as a member of Prof. YongKeun Park's research group. Now, he is a first author of six SCI and SCIE papers mainly on wave optics, Fourier optics, and quantitative phase imaging.

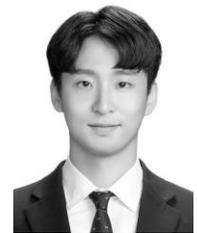
**Gunho Choi** is a researcher at Tomocube. He received the B.S. degree in Computer Science from Yonsei University, South Korea, in 2017. Previously, he worked at several start-up companies and developed deep learning models for classification, segmentation, and detection. His current research interest is to apply deep learning techniques to biomedical area.

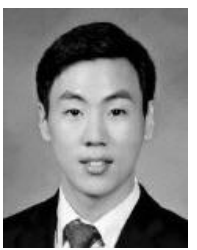
**Geon Kim** is a Ph.D. candidate in the Department of Physics at KAIST, South Korea. He received the B.S. degree in Physics from KAIST in 2016. He is currently affiliated with Prof. YongKeun Park's research group, where he has studied three-dimensional live cell imaging. His research interests include biomedical applications of microscopy techniques. Currently, he is primarily interested in learning-based approach to analyses of microscopic images.

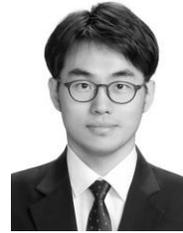
**Hyun-seok Min** is a researcher and the AI Team leader at Tomocube. He received the Ph.D. degree from KAIST, South Korea. Previously, he worked at Samsung Electronics, where he conducted research on various image processing areas including object recognition and image retrieval with machine learning algorithm. His current research interests include technologies of image processing, machine learning and deep learning for medical imaging problems.

**YongKeun (Paul) Park** is Associate Professor of Physics at KAIST, South Korea. He received the Ph.D. degree from 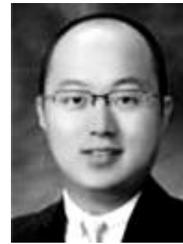 Harvard-MIT Health Sciences and Technology. Dr. Park's area of research is wave optics and its applications for biology and medicine. He has published +120 peer-reviewed papers including 3 Nat. Photon., 2 Nat. Comm., 4 PRL, 4 PNAS papers. Two start-up companies with +30 employees have been created from his research (Tomocube, The.Wave.Talk). Dr. Park's awards and honors include Fellow membership (Optical Society of America), Jinki Hong Creative Award, and Medal of Honor in Science and Technology (President of South Korea).